# Predictive Modeling and Uncertainty Quantification of Fatigue Life in Metal Alloys using Machine Learning

*Jiang Chang[1], Deekshith Basvoju[2], Aleksandar Vakanski[2*], Indrajit Charit[2], Min Xian[1]*

[1]Department of Computer Science, University of Idaho, Moscow, Idaho 83844, USA

[2]Department of Nuclear Engineering and Technology Management, University of Idaho, Idaho Falls, Idaho 83402, USA

**\*Correspondence:**
Aleksandar Vakanski: vakanski@uidaho.edu

## Abstract

Recent advancements in machine learning-based methods have demonstrated great potential for improved property prediction in material science. However, reliable estimation of the confidence intervals for the predicted values remains a challenge, due to the inherent complexities in material modeling. This study introduces a novel approach for uncertainty quantification in fatigue life prediction of metal materials based on integrating knowledge from physics-based fatigue life models and machine learning models. The proposed approach employs physics-based input features estimated using the Basquin fatigue model to augment the experimentally collected data of fatigue life. Furthermore, a physics-informed loss function that enforces boundary constraints for the estimated fatigue life of considered materials is introduced for the neural network models. Experimental validation on datasets comprising collected data from fatigue life tests for Titanium alloys and Carbon steel alloys demonstrates the effectiveness of the proposed approach. The synergy between physics-based models and data-driven models enhances the consistency in predicted values and improves uncertainty interval estimates. The codes are available at: https://github.com/avakanski/Fatigue-Life-Prediction

## 1     Introduction

The rapid design and development of advanced materials in material science engineering have been driven by three main paradigms: empirical, theoretical, and computational [1]–[3]. Accordingly, an enormous volume of theoretical and experimental data (TED) has been generated by these paradigms through methods such as Density Functional Theory and high-throughput approaches [4]. A crucial aspect of material development is modeling material properties, due to the impact on functionality, safety and reliability, quality control, and sustainability and environmental impact. Fatigue is among the most critical material properties, particularly in industries where material failure can have catastrophic consequences. Fatigue deformation represents a complex phenomenon that involves dynamic interactions between multi-physics and multi-scale processes [5]. Experimental evaluation of fatigue is time- and cost-intensive, because it requires conducting experiments involving large numbers of loading cycles and long periods to ensure the structural safety of materials. Conventional approaches for fatigue modeling typically employ physics-based models derived from knowledge of the governing physics laws and principles [6]. Although numerous physics-based models have been proposed in the literature, accurate mathematical formulation of the multi-physics and multi-scale complexities of fatigue deformation remains challenging, whereas physics-based models capture only the most important factors that influence fatigue and are missing fine details of the underlying physics [7,8].

Motivated by the recent success of ML approaches for pattern recognition in high-dimensional spatio-temporal data across various applications, a body of works employed ML models for predicting material properties as an alternative to physics-based models. Despite the promising capabilities of data-driven empirical models, two major obstacles require further attention from the research community [9]. First, very few datasets and benchmarks are publicly available (for fatigue modeling specifically), and of the available datasets, most are of limited size and/or lack consistent information regarding manufacturing processing methods, or other pertinent parameters. Training ML models with a small number of experimental data points necessitates extrapolating the model predictions into regions with sparsely populated values [10]. Consequently, although such ML predictions may be an excellent fit for the available experimental data, they may not truthfully reflect the modeled process. Second, the fatigue behavior of materials can be impacted by a multitude of multi-physics multi-scale phenomena, resulting in a high-dimensional parameter search space. This complexity can hamper the abilities of ML models trained solely on experimentally collected fatigue data to capture all underlying factors that affect fatigue [11]. As a result, ML predictions may not be consistent with the governing physics principles [12]. Conclusively, to harness the potential of ML for advancing fatigue modeling requires developing new computational tools that integrate analytical physics-based models and experimental data, and creating standardized benchmarks that offer large, structured datasets and implementation codes for method validation.

This work introduces a PIML approach for integrating UQ in predicting the fatigue life of metal alloys. Toward this goal, we study various ML models for multi-variable regression that provide uncertainty estimates, including the conventional ML models of quantile regression, natural gradient boosting, and Gaussian process regression, neural network (NN) models with deterministic parameters including deep ensembles and MC dropout, and Bayesian NNs with probabilistic parameters including variational inference networks and Markov chain Monte Carlo networks. The proposed PIML approach introduces physic-informed features for augmenting the fatigue datasets with estimated fatigue life based on the Basquin model, and we implemented a physics-informed loss function for the NN models that enforces boundary constraints in the predicted fatigue values. Experimental validation of the proposed approach demonstrates improved performance for fatigue datasets with Titanium and Carbon steel alloys.

The contributions of the proposed work in this paper are as follows:
- Introduced a PIML approach for integration of predicted fatigue life from physics-based models into machine learning models.
- Implemented regression methods for uncertainty quantification of fatigue life including conventional ML methods, NNs with deterministic, and Bayesian NNs with probabilistic parameters.
- Performed experimental evaluation with fatigue datasets including four types of metal alloys.

## 2 Related Works

PIML methods incorporate prior knowledge from established physical laws and principles into ML algorithms [13]. This integration enables PIML to generate estimates of target variables that are consistent with the known behavior of physical systems, avoiding inaccuracies that can occur when physical constraints are not considered [14]. PIML offers the potential to address the challenges associated with modeling material properties because it leverages the ability of ML methods to capture complex relationships in high-dimensional, multi-scale, and multi-physics data [15]. Furthermore, PIML is particularly suitable for addressing the challenges presented by sparse datasets in the domain of materials science. By leveraging known physical principles, PIML can enhance the performance where traditional data-driven models might fail [14]. The integration of experimentally collected historical data on material properties with physics-based models within the PIML framework is a promising approach for enhancing the long-term estimates of material behavior under different conditions [16].

Uncertainty quantification (UQ) is another important aspect of modeling material behavior that pertains to determining the level of confidence in the estimates obtained by empirical, theoretical, or computational models [17,18]. High levels of uncertainties in the estimated properties can lead to deviations of the exhibited practical material behavior from the project behavior by the model, potentially resulting in premature failure and catastrophic consequences in critical applications. Despite the utmost importance of UQ in materials science, many methods for modeling material properties output only the estimated value of the variables of interest (referred to as point estimates, or single-point estimates), without providing UQ via confidence intervals or other means. The development of data-driven ML methods for material properties that incorporate uncertainty estimates remains an important topic requiring further attention from the research community.

Prior works in the literature introduced PIML methods for estimating various properties of materials, including creep-fatigue life [19], creep fracture life [18], and tensile properties [20]. Several works proposed PIML methods for estimating the fatigue life of materials [21], including structural adhesive joints [10], adjustable bearings in rotary machines [22], and metal alloys [21,23,24]. Differently from previous related works, our proposed PIML framework for fatigue life estimation focuses on outputting both single-point estimates and uncertainty quantification of the fatigue life value. Furthermore, whereas most of the previous PIML approaches placed emphasis on integrating standard NNs with deterministic parameters and physics-based models, this study investigates the utility of PIML framework based on BNNs with stochastic parameters, which provide the means for UQ in the estimated fatigue life values.

## 3   Materials and Methods

This section first presents the methodological background related to physics-based fatigue models, followed by a brief overview of ML models for uncertainty quantification. Afterward, the proposed PIML approach is introduced, followed by a description of the materials and datasets used for experimental evaluation.

### 3.1   Physics-based Fatigue Methods

Fatigue is a process where cyclic loading leads to the initiation and propagation of cracks within a material. Cracks can occur even when the applied stress is significantly below the yield strength of the material [25,26]. Under cyclic loading, the material experiences a sequence of tension and compression phases, during which microscopic defects within the grain structure serve as the nucleation sites of fatigue cracks [27,28]. As the cyclic loading continues, the cracks grow traversing the material and when reach a critical size they can lead to catastrophic failure. Fatigue plays a significant role in the design of metal parts [25]. Given that engineering components frequently undergo cyclical loading, fatigue is responsible for approximately 90% of all mechanical failures [26].

**Stress-life methods**: This category of fatigue methods are also known as the S-N methods or Wöhler curve methods, and are commonly used to characterize the fatigue behavior of materials. The graphical representation under cyclic stress is based on stress amplitude ($S$) and number of cycles ($N$). Each point on the S-N curve represents a test result from a specimen [29]. The fatigue model is given as:

$$N_f = A(\Delta\sigma)^{-B} \tag{1}$$

where $N_f$ denotes the number of cycles to failure, $\Delta\sigma$ is the stress range, and $A$ and $B$ are material constants.

The relationship between the stress amplitude $\sigma$ and the number of cycles to failure $N_f$, typically represented on a log-log scale, is also known as the Basquin relation [30]:

$$\sigma = cN_f^m \tag{2}$$

Stromeyer [31] proposed a similar model based on a modified Basquin relation, which introduces the fatigue endurance limit of material $\sigma_f$, and it is given with:

$$\sigma = \sigma_f + cN_f^m \tag{3}$$

Another formulation related to the stress-life relationship is given by the Walker model [32], which considers the influence of the mean stress and the ratio of minimum to maximum stress. The Walker model defines an equivalent stress as:

$$\sigma_{eq} = \sigma_{max}^{1-\gamma}\sigma_a^{\gamma} = \sigma_{max}\left(\frac{1-R}{2}\right)^{\gamma} = \sigma_a\left(\frac{2}{1-R}\right)^{1-\gamma} \tag{4}$$

where $\sigma_{eq}$ denotes equivalent stress, $\sigma_{max}$ is maximum stress, $\sigma_a$ is the stress amplitude, $R$ is the stress ratio, and $\gamma$ is an empirically determined material constant (for metals ~0.5).

**Strain-life methods**: These methods are also known as $\varepsilon - N$ methods and are used to predict the lifespan of material under cyclic loading conditions. The $\varepsilon - N$ methods consider both the elastic and plastic deformation of a material. When stress is applied, strain accumulates as the material deforms. Over time, under cyclic stresses permanent deformation can occur, which can eventually lead to failure. At lower stress levels, materials typically experience elastic deformation, resulting in a linear relationship between stress and strain. At higher stress levels, the deformation causes permanent changes in the material structure leading to nonlinear relationship between stress and strain, as the material may exhibit yielding, hardening, or other behaviors [29,33]. The Coffin-Manson equation [36,37] of fatigue life incorporates terms for both elastic strain $\Delta\varepsilon_{el}$ in low-cycle fatigue regime and plastic strain $\Delta\varepsilon_{pl}$ in high-cycle fatigue regime to provide an improved fatigue model:

$$\Delta\varepsilon = \Delta\varepsilon_{el} + \Delta\varepsilon_{pl} = \frac{\sigma_{sc}}{E}(2N_f)^b + \varepsilon_f(2N_f)^c \tag{5}$$

where $\Delta\varepsilon$ denotes the strain range, $\sigma_{sc}$ is fatigue strength coefficient, $E$ is the modulus of elasticity, $\varepsilon_f$ is fatigue ductility coefficient, and $b$ and $c$ are fatigue strength exponent and fatigue ductility exponent, respectively.

**Crack Growth**: These methods in materials fatigue involve the study of how cracks in a material evolve under various loading conditions. The cracks typically form around pre-existing flaws in a part and grow during operational use.

SWT (Smith-Watson-Topper) model predicts the life of a component by summing up the width of each increment of crack growth for each loading cycle. The rate of crack growth is typically measured by applying constant amplitude cycles to a specimen and measuring the rate of growth from the change in compliance of the specimen or by measuring the growth of the crack on the surface of the specimen [36,37]. The model combines the maximum stress and the strain amplitude to predict fatigue life, and it is primarily applicable to low-cycle fatigue regimens. SWT model is given with:

$$\varepsilon_{SWT}\sigma_{max} = \varepsilon_a \sigma_a = \left(\frac{\sigma_{sc}}{E}(2N_f)^b + \varepsilon_f(2N_f)^c\right)\sigma_a \tag{6}$$

where $\varepsilon_{SWT}$ is SWT strain, $\sigma_{max}$ is the maximum tensile stress, $\varepsilon_a$ is the strain amplitude, $\sigma_a$ is the stress amplitude, $\sigma_{sc}$ is the fatigue strength coefficient, $E$ is the modulus of elasticity, and $\varepsilon_f$ is the fatigue ductility coefficient.

The critical plane approach uses the stress-strain parameter on the plane within the material where the maximum damage occurs, for predicting the initiation of cracks. When tensile cracking is the primary mode of crack initiation, the Smith-Watson-Topper (SWT) model identifies the critical plane as the plane with the maximum normal strain [38]. This critical plane is where tensile stresses are most likely to cause crack initiation, it is applicable to both low-cycle and high-cycle fatigue regimens, and the proposed model is [38]:

$$\sigma_{max}\frac{\Delta\varepsilon}{2} = \frac{\sigma_f^2}{E}(2N_f)^b + \sigma_{sc}\varepsilon_f(2N_f)^{b+c} \tag{7}$$

In the case of shear cracking, Fatemi and Socie [39] proposed the following model:

$$\frac{\Delta\gamma_{max}}{2}\left(1 + k\frac{\sigma_{n,max}}{\sigma_y}\right) = \frac{\tau_f'}{G}(2N_f)^{b_0} + \gamma_f'(2N_f)^{c_0} \tag{8}$$

where $\Delta\gamma_{max}$ denotes the maximum shear strain range on the critical plane, $\sigma_{n,max}$ is the maximum normal stress on the critical plane, $\sigma_y$ is the yield strength of the material, $G$ is the shear modulus, $k$ is a material constant that is fitted to uniaxial and torsional data, and $\tau_f'$, $b_0$, $\gamma_f'$, and $c_0$ are the fatigue parameters fitted by torsional data.

The model proposed by Xue et al. [40] uses the equivalent strain amplitude as the fatigue damage parameter. The parameter accounts for the non-proportional effect, and the critical plane is identified as the plane experiencing the maximum shear strain range. The model is expressed as

$$\left(\frac{\tau_{max}}{\tau_f'} + \frac{\sigma_{n,max}}{\sqrt{3}\sigma_f'}\right)\sqrt{3\varepsilon_n^{*2} + \left(\frac{\Delta\gamma_{max}}{2}\right)^2} = \frac{\tau_f'}{G}(2N_f)^{2b_0} + \gamma_f'(2N_f)^{b_0+c_0} \tag{9}$$

where $\varepsilon_n^*$ is the normal strain excursion between adjacent turning points of the maximum shear strain range on the critical plane. The energy-based models are considered to be applicable to both high-cycle and low-cycle fatigue regimens.

## 3.2 ML Regression Models for Uncertainty Quantification

ML-based approaches for material properties prediction are typically formulated as multi-variable regression problems for predicting a continuous target variable based on a set of input features. Specifically, the problem of fatigue life prediction can be solved by training an ML model that takes as input information related to the material type and composition, manufacturing and processing conditions, known properties of the material, and measurements from fatigue tests performed under specific conditions, and the objective is to predict the fatigue life expressed as the number of cycles to fracture. I.e., the set of $N$ observed data samples $\mathbf{X} = \{ \mathbf{x}_i | \mathbf{x}_i \in \mathbb{R}^d, i = 1, 2, \cdots, N \}$ and target values $\mathbf{Y} = \{ y_i | y_i \in \mathbb{R}, i = 1, 2, \cdots, N \}$ comprise the training dataset $\mathcal{D} = \{(\mathbf{x}_i, y_i)\}_{i=1}^{N}$. For a new data point $\mathbf{x}^*$ which does not belong to the observed dataset, and a mapping function $f$ of the trained model, the predicted target value is $y^* = f(\mathbf{x}^*)$, often referred to as single-point prediction or point estimate. Besides the point estimate, ML methods for UQ provide also a quantified measure of the variance of the predicted value $y^*$.

Several conventional ML models for regression tasks deliver uncertainty estimates of the predictions. *Quantile Regression* (QR) [41] is a non-parametric approach for estimating the conditional quantiles of predicted variables, whereas for a data point $\mathbf{x}^*$, the 95% prediction interval around the point estimate $y^*$ is obtained by calculating the conditional quantile function $Q_{y^*|\mathbf{x}^*}(\tau)$ for quantiles $\tau$ of 2.5% and 97.5%. *Natural Gradient Boosting* (NGBoost) regression [42] is a probabilistic variant of the traditional Gradient Boosting method [43]. For a data point $\mathbf{x}^*$, the point estimate of the target variable $y^*$ and the uncertainty estimate quantified as the standard deviation $\sigma^*$ are obtained from the conditional distribution $\mathcal{P}(y^*|\mathbf{x}^*, \theta)$. Advantages of NGBoost include the flexibility to be used with any other ML models as based learners. *Gaussian Process Regression* (GPR) [44] is a non-parametric Bayesian approach, that represents a collection of random variables as a multivariate Gaussian distribution over a set of data points. The smoothness of the distribution is determined by the covariance kernel $K_{i,j} = k(\mathbf{x}_i, \mathbf{x}_j)$ that defines the covariance between the function values $f(\mathbf{x}_i)$ and $f(\mathbf{x}_j)$ for data points $\mathbf{x}_i$ and $\mathbf{x}_j$. Given a training dataset $(\mathbf{X}, \mathbf{Y})$, for a new data point $\mathbf{x}^*$, the point and uncertainty estimates are obtained from the posterior distribution $\mathcal{P}(y^*|\mathbf{x}^*, \mathbf{X}, \mathbf{y}) = \mathcal{N}(\boldsymbol{\mu}^*, \boldsymbol{\sigma}^{*2})$. GPR is among the most powerful and flexible methods for UQ in regression tasks, where using different kernel functions and hyperparameters allows to introduce domain knowledge and adapt the predictive distribution to the specific dataset.

Neural Networks (NNs)-based methods have also been developed for UQ in multi-variable regression tasks. *Deep Ensembles* (DE) [45] employ the aggregated outputs of an ensemble of trained NNs to estimate the uncertainties in the target variable. For a new data point $\mathbf{x}^*$, the mean and standard deviation of the predictions by the DE are used as the point estimate and uncertainty estimate, respectively. *Monte Carlo (MC) Dropout* [46] utilizes a trained NN and applies dropout during inference to generate Monte Carlo samples. Similarly to DE, the resulting distribution of predicted values is used for calculating the point and uncertainty estimates.

Besides standard NNs with deterministic parameters, Bayesian Neural Networks (BNNs) have also been used for UQ, where the network parameters are represented with probability distributions [47–49]. For a BNN model parameterized with parameters $\theta$ that form probability distributions, inference for a new data point $\mathbf{x}^*$ is performed by using the posterior predictive distribution $\mathcal{P}(y^*|\mathbf{x}^*, \mathcal{D}) \propto \mathcal{P}(y^*|\mathbf{x}^*, \theta) \, \mathcal{P}(\theta|\mathcal{D})$. Direct calculation of the posterior distribution of the parameters given observed

data $\mathcal{P}(\theta|\mathcal{D})$ is intractable, and approximations are used in practice. *Variational Inference (VI) BNNs* employ an optimization technique to approximate the intractable posterior distribution $\mathcal{P}(\theta|\mathcal{D})$ with a simpler parameterized distribution $q_\phi(\theta)$ from a family of distributions $\mathcal{Q}$. The predictive uncertainty for new data point $\mathbf{x}^*$ is estimated by sampling from $q_\phi(\theta)$, which is referred to as variational distribution [42]. *Markov Chain Monte Carlo (MCMC) BNNs* approximate the posterior distribution of NN parameters $\theta$ given observational data through Monte Carlo sampling [50,51]. The approach employs a Markov Chain of model parameters, where each set of parameters is a sample from the posterior distribution. For new input data point $\boldsymbol{x}^*$, point and uncertainty estimates are calculated by averaging the predictions based on generated samples from $\mathcal{P}(\theta|\mathcal{D})$.

In this study, we applied conventional ML methods, standard NNs with deterministic parameters, and BNNs for predicting the fatigue life of metal alloys. The experimental results are presented in Section 4.

## 3.3 Physics-informed Machine Learning

Physics-Informed Machine Learning (PIML) integrates insights from the fundamental physics laws governing a process into ML models to improve the consistency of the predictions.

### 3.3.1 Physics-informed Feature Engineering

In our proposed PIML framework, we implemented physics-informed feature engineering, where the used datasets for predicting the fatigue life of metal alloys are augmented with physics-informed features. Specifically, we employed the Basquin model from equation (2) to estimate the fatigue life of an alloy for a given stress level based on experimental data from fatigue tests. We used the estimated values for fatigue life as additional input features for training the ML models.

### 3.3.2 Physics-informed Loss Function

Additionally, for the NN models, we implemented a physics-informed loss function that introduces constraints into the learning algorithm. Inspired by the work of Zhang et al. [19], we introduced two new loss terms to apply boundary conditions for the predicted fatigue life of metal alloys. The loss terms enforce the predicted fatigue life by the model to be non-negative and to be less than 10,000,000 cycles (which is considered in fatigue tests as the boundary for the endurance limit of the material). For ground-truth fatigue life value $y$ and predicted fatigue life $y^*$, the physics-informed loss is given with:

$$\mathcal{L} = \frac{1}{N}\sum_{i=1}^{N}[(y - y_i^*)^2 + \lambda_1 \text{ReLU}(y_i^*) + \lambda_2 \text{ReLU}(10{,}000{,}000 - y_i^*)] \tag{10}$$

The first loss term in (10) is the mean-squared error loss between the ground-truth and predicted fatigue values, the second term applies a ReLU activation function to impose non-negative predicted fatigue, and the third term similarly applies a ReLU activation function to impose predicted fatigue values less than 10,000,000 cycles. ReLU activation function is defined as $\text{ReLU}(x) = \{0 \text{ for } x < 0,\ x \text{ for } x \geq 0\}$. The weight coefficients $\lambda_1$ and $\lambda_2$ are used for balancing the impact of the loss terms on the model.

## 3.4 Datasets

In this study we utilized four fatigue datasets of metal alloys. The datasets contain information gathered from fatigue tests, and include details related to the chemical composition of the alloys, the manufacturing treatments and finishing processes applied to the test specimens, fatigue test conditions, and measured parameters from the fatigue tests. The collected data from the fatigue tests is employed in this work for training ML models, whereas the target variable is the fatigue life of the material expressed in number of cycles. These four datasets were selected because they contain information collected from hundreds of fatigue tests, providing a suitable basis for training and evaluating ML models.

The first fatigue dataset pertains to Titanium alloys and was adopted from Swetlana et al. [53]. It consists of 222 test samples with 24 features per sample. The features provide information about the weight percentage of the elements Ti, Al, V, Fe, C, H, O, Sn, Mb, Mo, Zr, Si, B, and Cr, testing conditions including applied stress (MPa) and temperature (°Celsius), finishing conditions related to the solution treated temperature (°Celsius), solution treated time (hours), annealing temperature (°Celsius), annealing time (hours), test measurements of the total strain (%), stress ratio, frequency, and the recorded fatigue life (number of cycles).

Three other fatigue datasets related to fatigue tests of Carbon steel alloys were extracted from the NIMS database [54]. The first dataset consists of 378 test samples from uniaxial tension-compression fatigue tests, the second dataset consists of 611 test samples from bending fatigue tests, and the third dataset has 208 test samples from torsion fatigue tests. The datasets have 18 features per sample, which provide the weight percent of the elements C, Si, Mn, P, S, Ni, Cr, and Cu, the test conditions including stress (MPa), temperature (°Celsius), reduction ratio (%), dA (inclusions due to plastic deformation), dB (discontinuously aligned inclusions), dC (isolated inclusions), elongation, area reduction, and the recorded fatigue life (number of cycles).

## 4 Results

### 4.1 Performance Metrics

The used metrics for point estimates include $R^2$ (coefficient of determination), PCC (Pearson correlation coefficient), RMSE (root-mean-squared error), and MAE (mean absolute error). Better performing models are characterized by high values of $R^2$ and PCC, and low values of RMSE and MAE, respectively. The employed metrics for uncertainty estimates include coverage (proportion of ground-truth values that fall within the predicted uncertainty interval), mean interval width (average size of the predicted uncertainty interval around the point estimates), and composite metric (adopted as $0.75 \cdot \text{coverage} + 0.25/\text{mean interval width}$). The motivation for introducing a composite metric is to combine the coverage and mean interval width into a single metric for UQ. Better preforming models are characterized with high values of coverage and composite metric, and low values of mean interval width.

### 4.2 Experimental Results

A high-level graphical representation of the proposed approach is shown in Figure 1. The fatigue dataset provides gathered information from fatigue tests, whereas the measured fatigue life of a material is considered a target variable, and the other parameters in the datasets are input features for the regression task. The proposed approach employs physics-informed feature engineering for estimating the fatigue life based on the provided information in the fatigue dataset. The input features in the dataset and the physics-informed fatigue life are used for training and evaluating ML models. For the NN models, a physics-informed loss function introduces constraints to the learning algorithm. The outputs of the ML models are predicted values of the fatigue life and uncertainty intervals.

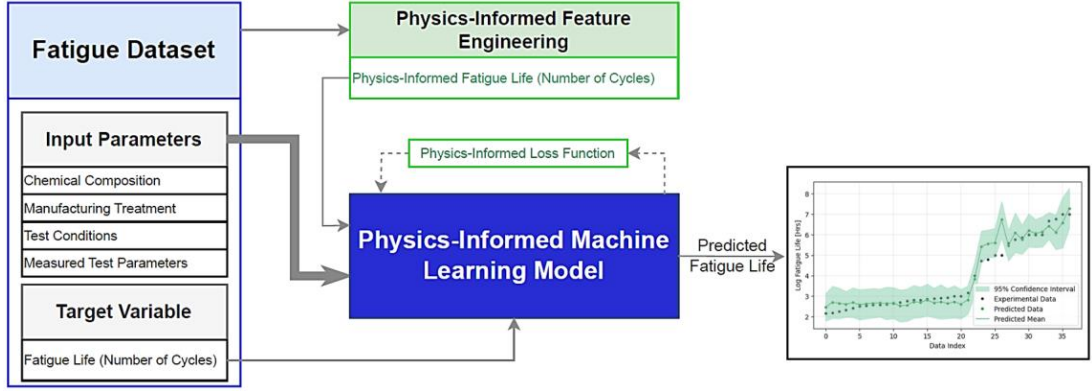

**Fig. 1.** High-level diagram of the proposed approach. The input parameters from a fatigue test dataset are augmented with physics-informed features and fed to a PIML model for predicting the fatigue life of a material.

The data pre-processing steps include normalizing the input features to a range between 0 and 1, and applying logarithm transformation to the fatigue life values. To evaluate the models 5-fold cross-validation was employed, that is, each run uses 80% of the data for training, and 20% for testing purposes. By applying the same random seed for all models, it was ensured that the folds contain the same data samples across the models. The mean and standard deviation of the metrics for the 5 folds are reported below.

The first experiment compares the performance of the studied regression models for the Titanium-alloy dataset. The results are presented in Table 1, where most models achieve high $R^2$ coefficient of determination, and over 0.9 Pearson correlation (PCC) between the predicted and ground-truth fatigue life for all models. BNN-MCMC model is the best performing with $R^2$ value of 0.9487, and PCC of 0.9751, which also has the lowest RMSE and MAE values. Deep Ensemble model also has good predicitive accuracy, however, the UQ for Deep Ensemble is unsatisfactory, where only 57.68% of the predicted values are within the confidence interval of 95%.

**Table 1.** Mean and standard deviation values for the Titanium alloys dataset.

| Model | $R^2$ ↑ | PCC ↑ | RMSE ↓ | MAE ↓ | Coverage Mean ↑ | Interval Width ↓ | Composite Metric ↑ |
|---|---|---|---|---|---|---|---|
| BNN-MCMC | **0.9487**$_{\pm 0.01}$ | **0.9751**$_{\pm 0.01}$ | **0.4148**$_{\pm 0.04}$ | **0.2921**$_{\pm 0.04}$ | 94.09$_{\pm 2.31}$ | 1.595$_{\pm 0.05}$ | 0.8626$_{\pm 0.01}$ |
| BNN-VI | 0.8835$_{\pm 0.02}$ | 0.9428$_{\pm 0.02}$ | 0.6367$_{\pm 0.06}$ | 0.4581$_{\pm 0.04}$ | 39.11$_{\pm 7.98}$ | 0.651$_{\pm 0.23}$ | 0.7773$_{\pm 0.30}$ |
| Deep Ensemble | 0.9086$_{\pm 0.01}$ | 0.9588$_{\pm 0.01}$ | 0.5658$_{\pm 0.03}$ | 0.3801$_{\pm 0.02}$ | 57.68$_{\pm 5.04}$ | 0.536$_{\pm 0.07}$ | **0.9092**$_{\pm 0.07}$ |
| GPR | 0.8778$_{\pm 0.04}$ | 0.9390$_{\pm 0.02}$ | 0.6490$_{\pm 0.10}$ | 0.4600$_{\pm 0.08}$ | 93.66$_{\pm 4.64}$ | 2.233$_{\pm 0.27}$ | 0.8159$_{\pm 0.03}$ |
| MC Dropout | 0.8928$_{\pm 0.02}$ | 0.9482$_{\pm 0.01}$ | 0.6100$_{\pm 0.06}$ | 0.4346$_{\pm 0.04}$ | 29.21$_{\pm 7.64}$ | **0.400**$_{\pm 0.09}$ | 0.8777$_{\pm 0.11}$ |
| NGBoost | 0.8814$_{\pm 0.02}$ | 0.9415$_{\pm 0.01}$ | 0.6443$_{\pm 0.05}$ | 0.4598$_{\pm 0.04}$ | 90.94$_{\pm 5.78}$ | 1.699$_{\pm 0.37}$ | 0.8370$_{\pm 0.02}$ |
| NN | 0.6960$_{\pm 0.35}$ | 0.9058$_{\pm 0.08}$ | 0.6443$_{\pm 0.05}$ | 2.0463$_{\pm 0.29}$ | - | - | - |
| QR | 0.8452$_{\pm 0.04}$ | 0.9244$_{\pm 0.02}$ | 0.7340$_{\pm 0.11}$ | 0.5124$_{\pm 0.08}$ | 81.05$_{\pm 4.95}$ | 1.890$_{\pm 0.87}$ | 0.7695$_{\pm 0.06}$ |

Table 2 presents the results for the Titanium alloy dataset by employing the PIML framework. The predictive accuracy for all models is enhanced in comparison to the results in Table 1, which is indicative of the efficacy of PIML approaches in refining the performance of the models. The best performing model is BNN-MCMC, with $R^2$ value of 0.9554 and $PCC$ of 0.9778. Graph with predicted values and uncertainty estimated by the models for one fold of the data is shown in Figure 2. In the figure, it is noticeable that models like Deep Ensemble and MC Dropout underestimate the uncertainties in the predictions. GPR provides both reliable point estimates and uncertainty estimates for the fatigue life.

**Table 2.** Mean and standard deviation values for the Titanium alloys dataset using the PIML method.

| Model | $R^2$ ↑ | PCC ↑ | RMSE ↓ | MAE ↓ | Coverage Mean ↑ | Interval Width ↓ | Composite Metric ↑ |
|---|---|---|---|---|---|---|---|
| BNN-MCMC | **0.9554**$_{\pm 0.012}$ | **0.9778**$_{\pm 0.063}$ | **0.3572**$_{\pm 0.051}$ | **0.2378**$_{\pm 0.028}$ | 92.97$_{\pm 4.045}$ | 1.1444$_{\pm 0.06}$ | 0.9163$_{\pm 0.02}$ |
| BNN-VI | 0.9439$_{\pm 0.010}$ | 0.9730$_{\pm 0.005}$ | 0.4440$_{\pm 0.054}$ | 0.3205$_{\pm 0.02}$ | 72.27$_{\pm 5.262}$ | 0.752$_{\pm 0.09}$ | 0.8797$_{\pm 0.02}$ |
| Deep Ensemble | 0.9505$_{\pm 0.016}$ | 0.9763$_{\pm 0.008}$ | 0.4153$_{\pm 0.076}$ | 0.2732$_{\pm 0.043}$ | 69.09$_{\pm 6.52}$ | 0.425$_{\pm 0.03}$ | 1.1094$_{\pm 0.04}$ |
| GPR | 0.9543$_{\pm 0.015}$ | 0.9776$_{\pm 0.008}$ | 0.3599$_{\pm 0.059}$ | 0.2219$_{\pm 0.028}$ | **98.81**$_{\pm 4.04}$ | 0.886$_{\pm 0.25}$ | 0.9811$_{\pm 0.08}$ |
| MC Dropout | 0.9464$_{\pm 0.019}$ | 0.9753$_{\pm 0.007}$ | 0.4282$_{\pm 0.081}$ | 0.2820$_{\pm 0.034}$ | 47.72$_{\pm 6.265}$ | **0.291**$_{\pm 0.044}$ | **1.2407**$_{\pm 0.134}$ |
| NG Boost | 0.9312$_{\pm 0.039}$ | 0.9656$_{\pm 0.02}$ | 0.4264$_{\pm 0.118}$ | 0.2792$_{\pm 0.074}$ | 89.18$_{\pm 7.047}$ | 0.9736$_{\pm 0.36}$ | 0.9816$_{\pm 0.13}$ |
| NN | 0.9019$_{\pm 0.037}$ | 0.9544$_{\pm 0.015}$ | 0.5814$_{\pm 0.12}$ | 2.0484$_{\pm 0.076}$ | - | - | - |
| QR | 0.9348$_{\pm 0.0185}$ | 0.9693$_{\pm 0.008}$ | 0.4320$_{\pm 0.073}$ | 0.2659$_{\pm 0.041}$ | 80.54$_{\pm 2.648}$ | 1.40$_{\pm 0.57}$ | 0.8117$_{\pm 0.074}$ |

The next tables present similar performance results for the Carbon Steel datasets. Specifically, Tables 3 and 4 present the results for the dataset obtained with uniaxial fatigue testing, Tables 5 and 6 present the results for the dataset obtained with bending fatigue testing, and Tables 7 and 8 present the results for the dataset obtained with torsion fatigue testing of Carbon Steel alloys. In all cases, the values of the performance metrics obtained with ML methods are first presented, followed by the corresponding values of the performance metrics obtained with PIML models. The experimental results in Tables 4, 6, and 8 show that the PIML approach improves the estimated fatigue life of all evaluated models in comparison to the results obtained with ML models from Tables 3, 5, and 7, respectively. Most of the metrics values for single point estimates have an improvement of approximately 10 to 20 percent after the use of the PIML method. For instance, the obtained $R^2$ value for BNN-MCMC in Table 5 of 0.7593 increased to 0.8519 in Table 6 when the PIML approach is employed.

Based on the results in Tables 3 to 8, the BNN-MCMC model achieved the best performance across almost all single-point estimate metrics, and it is the most promising method for UQ of fatigue life of metal alloys. Furthermore, GPR also provided robust single-point estimation performance and UQ capabilities, and overall its performance is competitive to BNN-MCMC. Although several models evinced excellent performance for single-point estimates, such as Deep Ensemble and BNN-VI, their performance for UQ was less satisfactory. For example, in Table 4, one can note that the best $R^2$ value is obtained by Deep Ensemble, however, the coverage of 60.73 indicates that the model underestimates the uncertainties in the fatigue life values. Similar patterns are observed for the experiments with Carbon Steel data based on bending and torsion fatigue tests.

**Table 3.** Mean and standard deviation values for the Carbon steel alloys with uniaxial fatigue test dataset.

| Model | $R^2$ ↑ | PCC ↑ | RMSE ↓ | MAE ↓ | Coverage Mean ↑ | Interval Width ↓ | Composite Metric ↑ |
|---|---|---|---|---|---|---|---|
| BNN-MCMC | 0.4506$_{\pm 0.03}$ | 0.6975$_{\pm 0.02}$ | 0.4483$_{\pm 0.09}$ | 0.3457$_{\pm 0.06}$ | 93.69$_{\pm 5.93}$ | 1.853$_{\pm 0.14}$ | 0.8382$_{\pm 0.04}$ |
| BNN-VI | 0.5154$_{\pm 0.02}$ | 0.7236$_{\pm 0.02}$ | 0.4311$_{\pm 0.06}$ | 0.3315$_{\pm 0.04}$ | 48.96$_{\pm 7.98}$ | 0.5322$_{\pm 0.23}$ | 0.8673$_{\pm 0.30}$ |
| Deep Ensemble | **0.6124**$_{\pm 0.01}$ | **0.8060**$_{\pm 0.01}$ | 0.3761$_{\pm 0.03}$ | **0.2704**$_{\pm 0.02}$ | 60.73$_{\pm 5.04}$ | 0.5199$_{\pm 0.07}$ | 0.9776$_{\pm 0.07}$ |
| GPR | 0.5508$_{\pm 0.04}$ | 0.7621$_{\pm 0.02}$ | 0.4036$_{\pm 0.10}$ | 0.2920$_{\pm 0.08}$ | 91.76$_{\pm 4.64}$ | 1.543$_{\pm 0.27}$ | 0.8579$_{\pm 0.03}$ |
| MC Dropout | 0.5410$_{\pm 0.02}$ | 0.7490$_{\pm 0.01}$ | 0.4096$_{\pm 0.06}$ | 0.3093$_{\pm 0.04}$ | 30.61$_{\pm 7.64}$ | **0.2945**$_{\pm 0.09}$ | **1.147**$_{\pm 0.11}$ |

| Model | $R^2 \uparrow$ | PCC $\uparrow$ | RMSE $\downarrow$ | MAE $\downarrow$ | Coverage Mean $\uparrow$ | Interval Width $\downarrow$ | Composite Metric $\uparrow$ |
|---|---|---|---|---|---|---|---|
| NG Boost | $0.5058_{\pm0.02}$ | $0.7268_{\pm0.01}$ | **$0.4352_{\pm0.05}$** | $0.3389_{\pm0.04}$ | **$94.18_{\pm5.78}$** | $1.60_{\pm0.37}$ | $0.869_{\pm0.02}$ |
| NN | $0.5664_{\pm0.35}$ | $0.8004_{\pm0.08}$ | $0.3982_{\pm0.05}$ | $0.5999_{\pm0.29}$ | - | - | - |
| QR | $0.4758_{\pm0.04}$ | $0.7213_{\pm0.02}$ | $0.4480_{\pm0.11}$ | $0.3460_{\pm0.08}$ | $76.32_{\pm4.95}$ | $1.479_{\pm0.87}$ | $0.7593_{\pm0.06}$ |

**Table 4** Mean and standard deviation Carbon steel alloys with uniaxial fatigue test dataset using the PIML method.

| Model | $R^2 \uparrow$ | PCC $\uparrow$ | RMSE $\downarrow$ | MAE $\downarrow$ | Coverage Mean $\uparrow$ | Interval Width $\downarrow$ | Composite Metric $\uparrow$ |
|---|---|---|---|---|---|---|---|
| BNN–MCMC | **$0.7612_{\pm0.03}$** | **$0.8763_{\pm0.02}$** | $0.3578_{\pm0.09}$ | $0.2844_{\pm0.06}$ | $95.76_{\pm5.93}$ | $1.462_{\pm0.14}$ | $0.8897_{\pm0.04}$ |
| BNN-VI | $0.7263_{\pm0.02}$ | $0.8584_{\pm0.02}$ | $0.3833_{\pm0.06}$ | $0.3001_{\pm0.04}$ | $50.54_{\pm7.98}$ | $0.503_{\pm0.23}$ | $0.8861_{\pm0.30}$ |
| Deep Ensemble | $0.7009_{\pm0.01}$ | $0.8487_{\pm0.01}$ | $0.4000_{\pm0.03}$ | $0.3065_{\pm0.02}$ | $46.81_{\pm5.04}$ | $0.494_{\pm0.07}$ | $0.9486_{\pm0.07}$ |
| GPR | $0.7560_{\pm0.04}$ | $0.8746_{\pm0.02}$ | $0.3611_{\pm0.10}$ | $0.2858_{\pm0.08}$ | $96.29_{\pm4.64}$ | $1.468_{\pm0.27}$ | $0.8928_{\pm0.03}$ |
| MC Dropout | $0.7251_{\pm0.02}$ | $0.8577_{\pm0.01}$ | $0.3841_{\pm0.06}$ | $0.3017_{\pm0.04}$ | $26.96_{\pm7.64}$ | **$0.2763_{\pm0.09}$** | **$1.1432_{\pm0.11}$** |
| NG Boost | $0.7372_{\pm0.02}$ | $0.8618_{\pm0.01}$ | $0.3764_{\pm0.05}$ | $0.2989_{\pm0.04}$ | $94.97_{\pm5.78}$ | $1.4639_{\pm0.37}$ | $0.8910_{\pm0.02}$ |
| NN | $0.5733_{\pm0.35}$ | $0.7924_{\pm0.08}$ | $0.4769_{\pm0.05}$ | $0.7447_{\pm0.29}$ | - | - | - |
| QR | $0.6907_{\pm0.04}$ | $0.8441_{\pm0.02}$ | $0.4077_{\pm0.11}$ | $0.3237_{\pm0.08}$ | $87.04_{\pm4.95}$ | $1.9553_{\pm0.87}$ | $0.7832_{\pm0.06}$ |

**Table 5.** Mean and standard deviation values for the Carbon steel alloys with bending fatigue test dataset.

| Model | $R^2 \uparrow$ | PCC $\uparrow$ | RMSE $\downarrow$ | MAE $\downarrow$ | Coverage Mean $\uparrow$ | Interval Width $\downarrow$ | Composite Metric $\uparrow$ |
|---|---|---|---|---|---|---|---|
| BNN-MCMC | $0.7593_{\pm0.03}$ | $0.8755_{\pm0.02}$ | $0.2956_{\pm0.09}$ | $0.2075_{\pm0.06}$ | **$93.78_{\pm5.93}$** | $1.0981_{\pm0.14}$ | $0.9322_{\pm0.04}$ |
| BNN-VI | $0.7490_{\pm0.02}$ | $0.8688_{\pm0.02}$ | $0.3000_{\pm0.06}$ | $0.2051_{\pm0.04}$ | $71.20_{\pm7.98}$ | $0.5069_{\pm0.23}$ | $1.0775_{\pm0.30}$ |
| Deep Ensemble | **$0.7654_{\pm0.01}$** | **$0.8838_{\pm0.01}$** | $0.2923_{\pm0.03}$ | $0.2048_{\pm0.02}$ | $66.77_{\pm5.04}$ | $0.4881_{\pm0.07}$ | $1.0954_{\pm0.07}$ |
| GPR | $0.7607_{\pm0.04}$ | $0.8751_{\pm0.02}$ | $0.2948_{\pm0.10}$ | $0.2049_{\pm0.08}$ | $93.29_{\pm4.64}$ | $1.1452_{\pm0.27}$ | $0.9212_{\pm0.03}$ |
| MC Dropout | $0.7647_{\pm0.02}$ | $0.8815_{\pm0.01}$ | $0.2924_{\pm0.06}$ | **$0.2030_{\pm0.04}$** | $50.57_{\pm7.64}$ | **$0.3248_{\pm0.09}$** | **$1.2478_{\pm0.11}$** |
| NG Boost | $0.5309_{\pm0.02}$ | $0.7749_{\pm0.01}$ | $0.4136_{\pm0.05}$ | $0.2937_{\pm0.04}$ | $92.96_{\pm5.78}$ | $1.4574_{\pm0.37}$ | $0.8768_{\pm0.02}$ |
| NN | $0.6486_{\pm0.35}$ | $0.8386_{\pm0.08}$ | $0.3589_{\pm0.05}$ | $0.5831_{\pm0.29}$ | - | - | - |
| QR | $0.7089_{\pm0.04}$ | $0.8513_{\pm0.02}$ | $0.3245_{\pm0.11}$ | $0.2148_{\pm0.08}$ | $88.21_{\pm4.95}$ | $1.5903_{\pm0.87}$ | $0.8295_{\pm0.06}$ |

**Table 6.** Mean and standard deviation values for the Carbon steel alloys with bending fatigue test dataset using the PIML method.

| Model | $R^2 \uparrow$ | PCC $\uparrow$ | RMSE $\downarrow$ | MAE $\downarrow$ | Coverage Mean $\uparrow$ | Interval Width $\downarrow$ | Composite Metric $\uparrow$ |
|---|---|---|---|---|---|---|---|
| BNN-MCMC | **$0.8519_{\pm0.03}$** | **$0.9244_{\pm0.02}$** | $0.2322_{\pm0.09}$ | $0.1640_{\pm0.06}$ | **$93.95_{\pm5.93}$** | $0.8417_{\pm0.14}$ | $1.0058_{\pm0.04}$ |
| BNN-VI | $0.8411_{\pm0.02}$ | $0.9180_{\pm0.02}$ | $0.2382_{\pm0.06}$ | $0.1697_{\pm0.04}$ | $68.91_{\pm7.98}$ | $0.4016_{\pm0.23}$ | $1.1995_{\pm0.30}$ |
| Deep Ensemble | $0.8307_{\pm0.01}$ | $0.9155_{\pm0.01}$ | $0.2487_{\pm0.03}$ | $0.1733_{\pm0.02}$ | $79.22_{\pm5.04}$ | $0.3006_{\pm0.07}$ | $1.2825_{\pm0.07}$ |
| GPR | $0.8392_{\pm0.04}$ | $0.9169_{\pm0.02}$ | $0.2421_{\pm0.10}$ | $0.1767_{\pm0.08}$ | $92.80_{\pm4.64}$ | $0.9536_{\pm0.27}$ | $0.9601_{\pm0.03}$ |
| MC Dropout | $0.8075_{\pm0.02}$ | $0.9034_{\pm0.01}$ | $0.2654_{\pm0.06}$ | $0.1898_{\pm0.04}$ | $38.61_{\pm7.64}$ | **$0.2187_{\pm0.09}$** | **$1.5018_{\pm0.11}$** |
| NG Boost | $0.8372_{\pm0.02}$ | $0.9172_{\pm0.01}$ | $0.2426_{\pm0.05}$ | $0.1660_{\pm0.04}$ | $92.29_{\pm5.78}$ | $0.7897_{\pm0.37}$ | $1.0470_{\pm0.02}$ |
| NN | $0.6429_{\pm0.35}$ | $0.8296_{\pm0.08}$ | $0.3608_{\pm0.05}$ | $0.5833_{\pm0.29}$ | - | - | - |
| QR | $0.7836_{\pm0.04}$ | $0.8948_{\pm0.02}$ | $0.2803_{\pm0.11}$ | $0.1839_{\pm0.08}$ | $91.15_{\pm4.95}$ | $1.2966_{\pm0.87}$ | $0.8515_{\pm0.06}$ |

**Table 7.** Mean and standard deviation values for the Carbon steel alloys with torsion fatigue test dataset.

| Model | $R^2$ ↑ | PCC ↑ | RMSE ↓ | MAE ↓ | Coverage Mean ↑ | Interval Width ↓ | Composite Metric ↑ |
|---|---|---|---|---|---|---|---|
| BNN-MCMC | $0.6414_{\pm 0.03}$ | $0.8129_{\pm 0.02}$ | $0.4370_{\pm 0.09}$ | $0.3460_{\pm 0.06}$ | $95.50_{\pm 5.93}$ | $1.7434_{\pm 0.14}$ | $0.8596_{\pm 0.04}$ |
| BNN-VI | $0.6386_{\pm 0.02}$ | $0.8046_{\pm 0.02}$ | $0.4407_{\pm 0.06}$ | $0.3424_{\pm 0.04}$ | $59.00_{\pm 7.98}$ | $0.6688_{\pm 0.23}$ | $0.8154_{\pm 0.30}$ |
| Deep Ensemble | $0.6055_{\pm 0.01}$ | $0.7983_{\pm 0.01}$ | $0.4566_{\pm 0.03}$ | $\mathbf{0.3515}_{\pm 0.02}$ | $47.88_{\pm 5.04}$ | $0.5407_{\pm 0.07}$ | $0.8698_{\pm 0.07}$ |
| GPR | $0.6495_{\pm 0.04}$ | $\mathbf{0.8171}_{\pm 0.02}$ | $0.4319_{\pm 0.10}$ | $0.3445_{\pm 0.08}$ | $\mathbf{96.03}_{\pm 4.64}$ | $1.7791_{\pm 0.27}$ | $0.8619_{\pm 0.03}$ |
| MC Dropout | $\mathbf{0.6501}_{\pm 0.02}$ | $\mathbf{0.8171}_{\pm 0.01}$ | $\mathbf{0.4305}_{\pm 0.06}$ | $0.3391_{\pm 0.04}$ | $37.83_{\pm 7.64}$ | $\mathbf{0.4465}_{\pm 0.09}$ | $\mathbf{0.8980}_{\pm 0.11}$ |
| NG Boost | $0.2367_{\pm 0.02}$ | $0.5471_{\pm 0.01}$ | $0.6439_{\pm 0.05}$ | $0.5227_{\pm 0.04}$ | $96.02_{\pm 5.78}$ | $2.404_{\pm 0.37}$ | $0.8256_{\pm 0.02}$ |
| NN | $0.5255_{\pm 0.35}$ | $0.7742_{\pm 0.08}$ | $0.5013_{\pm 0.05}$ | $0.7418_{\pm 0.29}$ | - | - | - |
| QR | $0.3541_{\pm 0.04}$ | $0.6316_{\pm 0.02}$ | $0.5912_{\pm 0.11}$ | $0.4743_{\pm 0.08}$ | $88.90_{\pm 4.95}$ | $2.3467_{\pm 0.87}$ | $0.7730_{\pm 0.06}$ |

**Table 8.** Mean and standard deviation values for the Carbon steel alloys with torsion fatigue test dataset using the PIML method.

| Model | $R^2$ ↑ | PCC ↑ | RMSE ↓ | MAE ↓ | Coverage Mean ↑ | Interval Width ↓ | Composite Metric ↑ |
|---|---|---|---|---|---|---|---|
| BNN-MCMC | $\mathbf{0.7612}_{\pm 0.03}$ | $\mathbf{0.8763}_{\pm 0.02}$ | $\mathbf{0.3578}_{\pm 0.09}$ | $\mathbf{0.2844}_{\pm 0.06}$ | $95.76_{\pm 5.93}$ | $1.4620_{\pm 0.14}$ | $0.8897_{\pm 0.04}$ |
| BNN-VI | $0.7263_{\pm 0.02}$ | $0.8584_{\pm 0.02}$ | $0.3833_{\pm 0.06}$ | $0.3001_{\pm 0.04}$ | $50.54_{\pm 7.98}$ | $0.5032_{\pm 0.23}$ | $0.8861_{\pm 0.30}$ |
| Deep Ensemble | $0.7009_{\pm 0.01}$ | $0.8487_{\pm 0.01}$ | $0.4000_{\pm 0.03}$ | $0.3065_{\pm 0.02}$ | $46.81_{\pm 5.04}$ | $0.4942_{\pm 0.07}$ | $0.9486_{\pm 0.07}$ |
| GPR | $0.7560_{\pm 0.04}$ | $0.8746_{\pm 0.02}$ | $0.3611_{\pm 0.10}$ | $0.2858_{\pm 0.08}$ | $\mathbf{96.29}_{\pm 4.64}$ | $1.4688_{\pm 0.27}$ | $0.8928_{\pm 0.03}$ |
| MC Dropout | $0.7251_{\pm 0.02}$ | $0.8577_{\pm 0.01}$ | $0.3841_{\pm 0.06}$ | $0.3017_{\pm 0.04}$ | $26.96_{\pm 7.64}$ | $\mathbf{0.2763}_{\pm 0.09}$ | $\mathbf{1.1432}_{\pm 0.11}$ |
| NG Boost | $0.7372_{\pm 0.02}$ | $0.8618_{\pm 0.01}$ | $0.3764_{\pm 0.05}$ | $0.2989_{\pm 0.04}$ | $94.97_{\pm 5.78}$ | $1.4639_{\pm 0.37}$ | $0.8910_{\pm 0.02}$ |
| NN | $0.5733_{\pm 0.35}$ | $0.7924_{\pm 0.08}$ | $0.4769_{\pm 0.05}$ | $0.7447_{\pm 0.29}$ | - | - | - |
| QR | $0.6907_{\pm 0.04}$ | $0.8441_{\pm 0.02}$ | $0.4077_{\pm 0.11}$ | $0.3237_{\pm 0.08}$ | $87.04_{\pm 4.95}$ | $1.9553_{\pm 0.87}$ | $0.7832_{\pm 0.06}$ |

Figure 3 presents the experimental fatigue life and the estimated fatigue life by the BNN-MCMC method for one fold of all four datasets. The subfigures (a), (c), (e), and (g) depict the obtained results with the original BNN-MCMC model, and the subfigures (b), (d), (f), and (h) depict the obtained results with the PIML version of the BNN-MCMC model. The subfigures (a) and (b) correspond to the Titanium alloy dataset, subfigures (c) and (d) correspond to the Carbon Steel alloy dataset with uniaxial tension-compression fatigue test, subfigures (e) and (f) correspond to the Carbon Steel alloy dataset with bending fatigue test, and subfigures (g) and (h) correspond to the Carbon Steel alloy dataset with torsion fatigue test. Importantly, Figure 3 visualizes the estimated confidence intervals by the models. One can observe in the figure that the calculated fatigue life by the PIML models is more accurate than the original models without the PIML approach, with the PIML estimated value having less deviation from the ground-truth values obtained from the actual fatigue tests. Additionally, the confidence intervals by the PIML models have higher coverage of the ground-truth fatigue life values, as well as they exhibit lower levels of overestimation of the uncertainties in the fatigue life values.

### 4.3 Implementation Details

For the QR, NGBoost, and GPR models we adopted hyperparameters from a previous related study for UQ in creep life prediction by Mamun et al. [55]. Model training and evaluation were performed using the scikit-learn, ngboost, and catboost libraries.

For the Deep Ensemble model, we used 5 base learners each comprising a standard NN with three fully-connected layers containing 10 neurons followed by dropout layers with a rate of 0.5 and ReLU activation. For the MC Dropout model, we used a standard NN with the same architecture, i.e., with three fully-connected layers containing 10 neurons followed by dropout layers with a rate of 0.5 and ReLU activation. For standard NN with deterministic parameters, we used three fully-connected layers with 1000, 200, and 40 neurons respectively, with ReLU activation function. For model training, we used MSE loss and Adam optimizer with a learning rate of 0.01. The experiment results were computed as the mean and ±3 standard deviations of the predictions by the base learners for Deep Ensemble, and based on generated 1,000 predictions for MC Dropout. We used the PyTorch library for implementing Deep Ensemble, MC Dropout, and standard NNs.

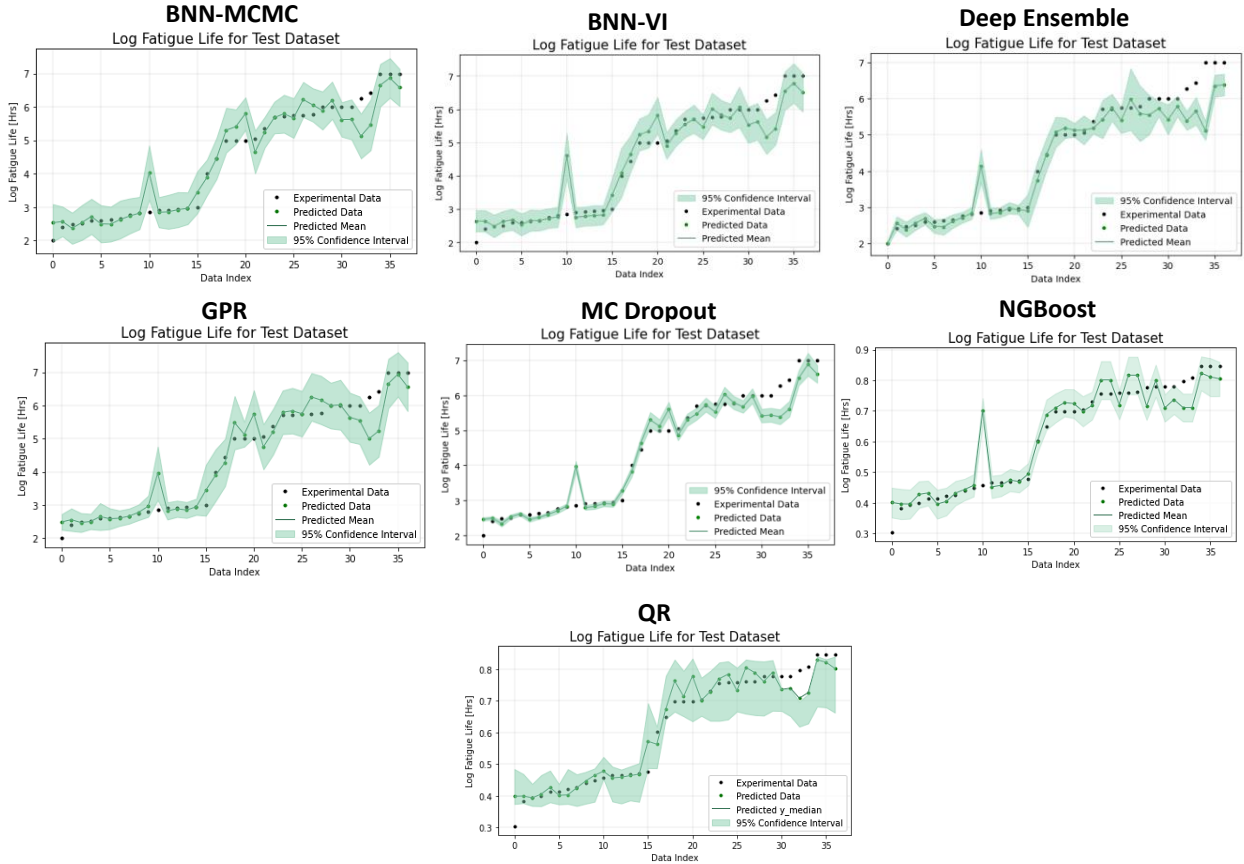

**Fig. 2.** Experimental data points, predicted data points, and uncertainty estimates using the PIML method for the Titanium alloy dataset. The green shaded area represents the 95% confidence interval for the predictions by the models.

For BNN-VI, we used an architecture comprised of two fully-connected layers with 100 neurons and ReLU activation functions. We employed a normal distribution with a mean of 0 and a standard deviation of 0.06 as the prior distribution for network parameters. For training, we utilized Stochastic Gradient Descent (SGD) with Nesterov Momentum and a learning rate of 0.001. Inference involved generating 1,000 samples from the variational distribution to calculate point estimates and uncertainty estimates. For BNN-MCMC, we employed an architecture with five fully-connected layers containing 10 neurons. Network parameter priors followed a normal distribution with a mean of 0 and a standard deviation of 1. For approximation of the posterior distribution we used the No-U-Turn Sampling (NUTS) algorithm. Inference involved drawing 100 samples with point and uncertainty estimates. We used the torchbnn library for BNN-VI and the Pyro library for implementing BNN-MCMC.

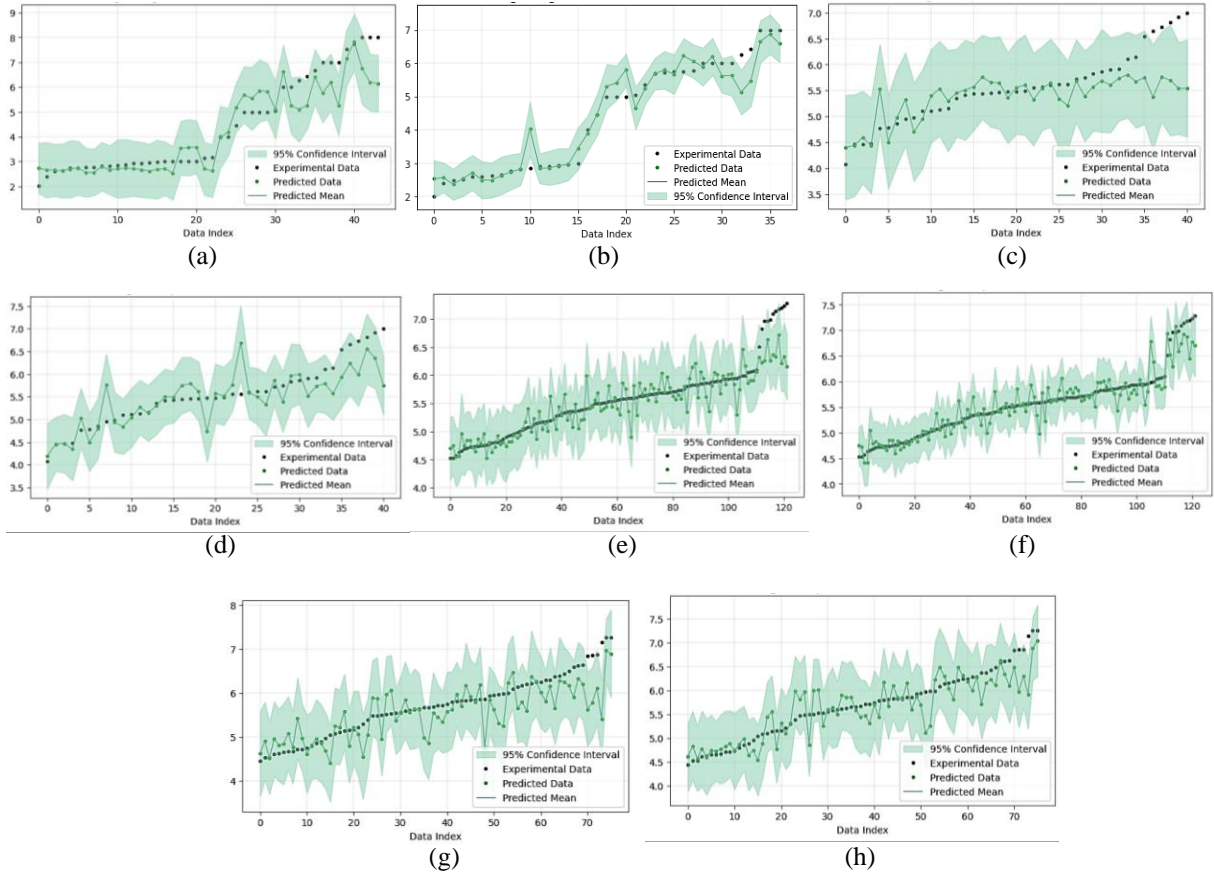

**Fig. 3.** Experimental fatigue life, estimated fatigue life, and uncertainty estimates obtained with the BNN-MCMC method for one fold of the datasets. (a) and (b) Titanium dataset; (c) and (d) Carbon Steel dataset uniaxial fatigue; (e) and (f) Carbon Steel dataset bending fatigue; (g) and (h) Carbon Steel dataset torsion fatigue. For all datasets, the first figure represents ML model and the second figure represents the PIML model. The green shaded area represents the 95% confidence interval for the estimates by the models.

## 5   Discussion

The experimental results highlight the potential of PIML methods for estimating the fatigue life of metal alloys and for uncertainty quantification. For all four datasets of fatigue tests, the models that leveraged physics-based information to guide the learning process achieved improved performance when compared to the models that were solely data-driven. Additionally, among the evaluated models, BNN-MCMC holds the greatest potential for accurate and reliable single-point and uncertainty estimates. The performance of GPR was also comparable and competitive to BNN-MCMC. Although the single-point estimates of fatigue life by the other evaluated models were satisfactory, these models exhibited overconfidence or underconfidence in uncertainty estimates. On the other hand, a limitation of BNNs-based models is their computational expense in comparison to GPR and conventional NNs and ML methods, since they require sampling from the posterior distribution and take a large number of sampling iterations to convergence. The need for increased computational resources can be taken into consideration when selecting the preferred approach for a specific task.

The findings of this study can be employed for the development and qualification of new materials. Specifically, estimated values of fatigue life and uncertainty quantification can be used in designing fatigue tests for newly developed materials, where they can reduce the experimental efforts by decreasing the number of required fatigue tests or the number of cycles of the tests. Similarly, the estimates can be employed for the design of test conditions for performing fatigue tests for new materials, as well as for the accelerated development and qualification of advanced materials with novel compositions.

# 6     Conclusion

This paper proposes an ML approach for estimating the fatigue life of metal alloys, with an emphasis on uncertainty quantification. The evaluated models for UQ include the conventional ML methods of quantile regression, natural gradient boosting, Gaussian process regression. Furthermore, various deep learning-based models were evaluated, encompassing NNs with deterministic parameters including deep ensembles and MC dropout, as well as Bayesian NNs with probabilistic parameters based on variational inference and Markov chain Monte Carlo networks. The proposed PIML approach adds additional features to the experimental data calculated based on the Basquin fatigue model. In addition, a PIML loss was introduced in the NN models with boundary constraints about the fatigue life. Experimental validation with datasets of Titanium alloys and Carbon steel alloys evinces improved accuracy of the PIML approach. Among the most reliable methods for UQ in fatigue life estimation are BNN-MCMC and GPR.